\documentclass{article}

\usepackage{amsmath}

\usepackage[final]{neurips_2024}

\usepackage[utf8]{inputenc} % allow utf-8 input
\usepackage[T1]{fontenc}    % use 8-bit T1 fonts
\usepackage{hyperref}       % hyperlinks
\usepackage{url}            % simple URL typesetting
\usepackage{booktabs}       % professional-quality tables
\usepackage{amsfonts}       % blackboard math symbols
\usepackage{nicefrac}       % compact symbols for 1/2, etc.
\usepackage{microtype}      % microtypography
\usepackage{xcolor}         % colors
\usepackage{subfigure}
\usepackage{multirow}
\usepackage{graphicx}
\usepackage{bm}

\usepackage{enumitem}

\title{Improving Equivariant Model Training via Constraint Relaxation}

\author{%
    Stefanos Pertigkiozoglou\thanks{Equal Contribution}\\
    University of Pennsylvania\\
    \texttt{pstefano@seas.upenn.edu}\\
    \And 
    Evangelos Chatzipantazis\footnotemark[1] \\
    University of Pennsylvania\\
    \texttt{vaghat@seas.upenn.edu}\\
    \And 
    Shubhendu Trivedi \\
    Independent \\
    \texttt{shubhendu@csail.mit.edu}
    \And
    Kostas Daniilidis \\
    University of Pennsylvania \\
    Archimedes, Athena RC\\
    \texttt{kostas@cis.upenn.edu}\\
}

\begin{document}

\maketitle

\begin{abstract}
Equivariant neural networks have been widely used in a variety of applications due to their ability to generalize well in tasks where the underlying data symmetries are known. Despite their successes, such networks can be difficult to optimize and require careful hyperparameter tuning to train successfully. In this work, we propose a novel framework for improving the optimization of such models by relaxing the hard equivariance constraint \emph{during} training: We relax the equivariance constraint of the network's intermediate layers by introducing an additional non-equivariant term that we progressively constrain until we arrive at an equivariant solution. By controlling the magnitude of the activation of the additional relaxation term, we allow the model to optimize over a larger hypothesis space containing approximate equivariant networks and converge back to an equivariant solution at the end of training. We provide experimental results on different state-of-the-art network architectures, demonstrating how this training framework can result in equivariant models with improved generalization performance. Our code is available at \href{https://github.com/StefanosPert/Equivariant_Optimization_CR}{https://github.com/StefanosPert/Equivariant\_Optimization\_CR}
\end{abstract}

\section{Introduction}\label{sec:intro}

The explicit incorporation of task-specific symmetry in the design and implementation of effective and parameter-efficient neural network (NN) models has matured into a rational and attractive NN design meta-formalism in recent years---that of group equivariant convolutional neural networks (GCNNs)~\citep{Cohen2016GroupEC, RavanbakhshSP17, esteves2018learning, Kondor2018OnTG, cohenGeneralTheoryEquivariant2019, Maron2019InvariantAE, e2cnn, Bekkers20, villar2021scalars, XuLDD22, pearce2023brauer}. GCNNs involve using the machinery of group and representation theory to compose layers that are equivariant to transformations of the input. Such networks, with hard-coded symmetry, have proven to be successful across a wide variety of tasks, while often affording significant data efficiency. Such tasks/domains include: RNA structure~\citep{Townshend2021GeometricDL}, protein structure~\citep{Baek2021AccuratePO, jumper2021highly}, molecule generation~\citep{Satorras2021EnEN}, medical imaging~\citep{Winkels2019PulmonaryND}, natural language processing~\citep{Gordon2020PermutationEM, petrache2024position}, computer vision~\citep{chatzipantazis2023mathrmseequivariant}, robotics~\citep{ZhuWangGrasp2022, ordonez2024morphological}, density functional theory~\citep{gong2023general}, particle physics~\citep{bogatskiy2020lorentz}, lattice gauge theories~\citep{boyda2021sampling} amongst many others. GCNNs now have also matured enough to have a well-developed theory. This includes both prescriptive (or architectural) theory and descriptive analysis. In general, GCNNs particularly stand out in domains with data scarcity, or with a high degree of symmetry~\citep{KufelApproximatelyinvariantNN,boyda2021sampling}, or in the physical sciences where respecting explicit symmetries could be dictated by physical laws, violating which could lead to physically implausible predictions.

Despite the successes of group equivariant models, there are several outstanding challenges that don't yet have general satisfactory solutions. We discuss two that have attracted recent attention. The first challenge---the primary motivation of our paper---has to do with the common observation that equivariant networks can be difficult to train~\citep{wang2024discoveringsymmetrybreakingphysical, kondor2018clebsch, liao2023equiformer}. The reasons for this general difficulty are not well-understood, but it seems to occur in part because the training dynamics of such networks can be notably different from non-equivariant architectures. For instance, if a GCNN operates entirely in Fourier space~\citep{bogatskiy2020lorentz, kondor2018clebsch, XuLDD22}, most of the usual intuition about training NN models does not apply. Further, depending on the level of equivariance error tolerance for a task, the internal layers could be computationally intensive, and involve e.g. higher-order tensor products. Notably, the above difficulty arises even when the model is correctly specified i.e. the model and the data encode the same symmetry. The second challenge with GCNNs, has to do with the downsides of working with exact equivariance when the data itself might have some (possibly) relaxed symmetry. This has recently led to a spurt of work on developing more flexible networks that can vary the amount of equivariance depending on the task~\citep{finzi2021residual, romero2022learning,ouderaa2022relaxing, wang2022approximately, Huang2023ApproximatelyEG, PetracheApproximate}. Such models generally improve accuracy and will sometimes also simplify the optimization process as a side-effect. Broadly, proposed solutions involve adding additional regularization terms that penalize for relaxation errors, solving for the problem of model mis-specification~\citep{PetracheApproximate}.

However, even though there is now work on relaxed\footnote{We use ``relaxed'' to encompass notions of partial and approximate equivariance~\citep{PetracheApproximate}.} equivariant networks that addresses model mis-specification, existing works don't focus on improving the optimization process directly. In this paper, we take a step towards examining this question in more detail. We make the case that \emph{even if we assume that the model is correctly specified}, relaxing the equivariance constraint during optimization and then projecting back to the equivariant space can itself help in improving performance. We conjecture that a prime reason for the optimization difficulty of GCNNs, as compared to non-equivariant models, is that their solution-space might be too severely constrained. We derive regularization terms that encourage each layer to operate in a larger hypothesis space during training---than being constrained to only be in the intertwiner space---while encouraging equivariant solutions. After the optimization is complete, we project the solution back onto the space of equivariant solutions. The approach can also be adapted to better optimize approximately equivariant networks in a similar manner. The focus of our work thus distinguishes it from works on relaxed equivariance---we are not concerned with mis-specification, but rather with isolating the optimization process itself.

\begin{figure}
    \centering
    \includegraphics[width=0.9\textwidth]{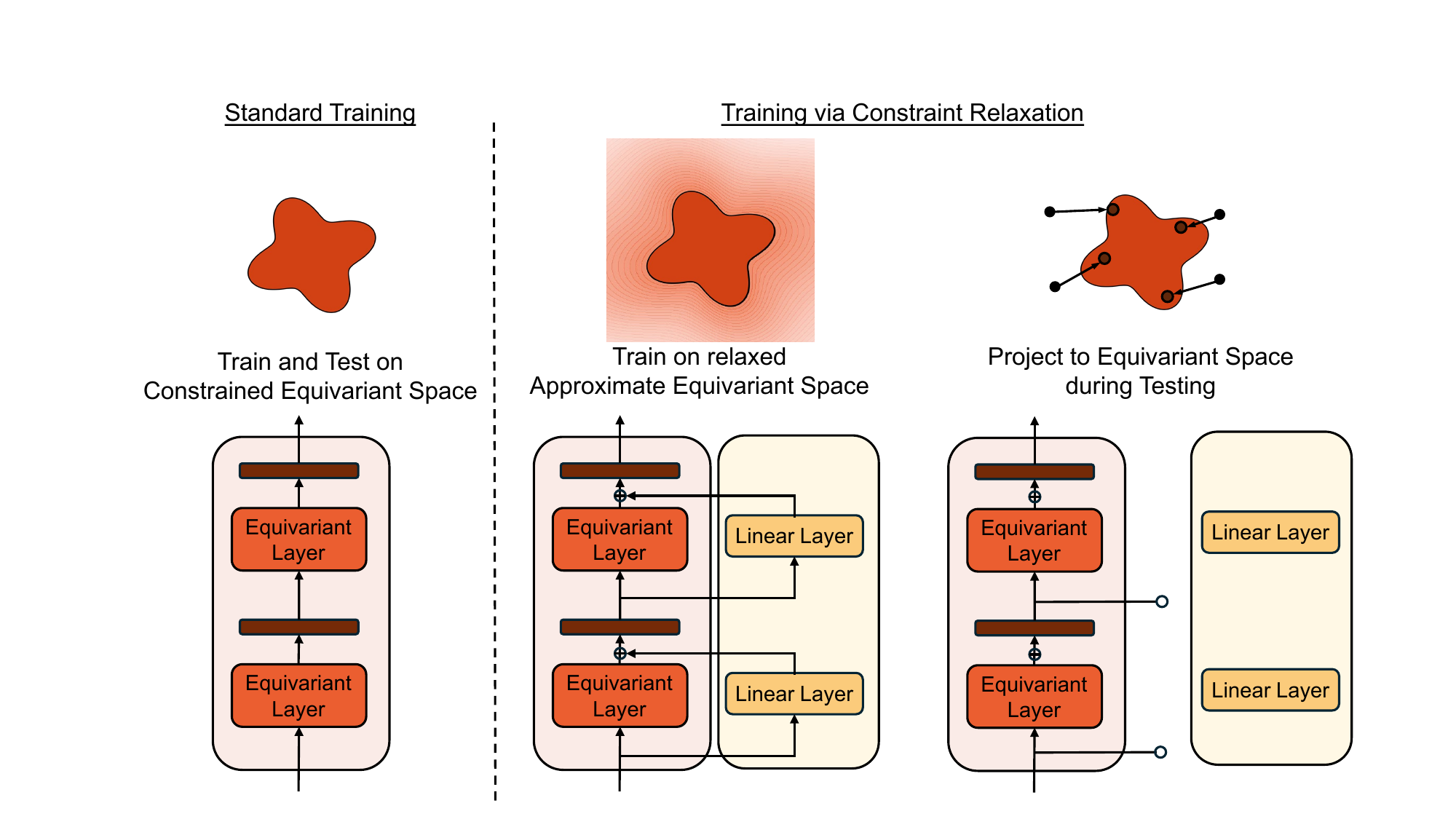}
    \caption{Standard training of equivariant NNs is constrained to a limited parameter space which can result in a challenging training process. We propose to relax these equivariant constraints during training, allowing optimization over a broader space of approximately equivariant models. During testing, we project the trained model back to the constrained space---arriving at an equivariant model with enhanced performance compared to equivalent models trained with the standard process.}
    \label{fig:method}
\end{figure}

Below we summarize the main contributions of our work:
\begin{itemize}[noitemsep]
    \item We present a novel training framework that can improve the performance of equivariant neural networks by relaxing the equivariance constraint during training and projecting back to the space of equivariant models during testing (as shown in Figure \ref{fig:method}).
    \item We present a formulation that extends existing equivariant neural network architectures to be approximately equivariant. We show how training on the relaxed network can improve the performance of its equivariant subnetwork.
    \item We provide experimental evidence showcasing how our framework improves the performance of existing state-of-the-art equivariant architectures.
\end{itemize}
\section{Related Work}\label{sec:related}
There is little prior work on providing general procedures for improving the optimization process for equivariant neural networks directly. \citet{elesedy2021provably} sketched a projected gradient method to construct equivariant networks and suggested a regularization scheme that could be used to implement approximate equivariance. However, this was proposed as an aside in the paper (sections 7.2 and 7.3), without empirical or theoretical backing. Our work also involves a projected gradient procedure. However, the regularization scheme that we propose is substantially different. 

Work on approximate and partial equivariance~\citep{finzi2021residual, romero2022learning,ouderaa2022relaxing, wang2022approximately, Huang2023ApproximatelyEG, PetracheApproximate,wang2023relaxed} seems superficially related to ours, but comes with a different motivation. Such methods aim to match data symmetry with model symmetry and are not explicitly concerned with improving optimization. As a result, they are designed to address tasks with either inherent relaxed symmetries or tasks where the underlying relaxed symmetry is misspecified. Contrary to that, our method focuses on cases where the underlying symmetry is known exactly, and the relaxation of the equivariant constraint is used only during training as a way to improve the optimization.  The works of~\cite{finzi2021residual,ouderaa2022relaxing, gruver2023the, otto2024unifiedframe, PetracheApproximate} are nonetheless relevant since they provide methods for measuring relaxed equivariance, comprising of regularization schemes that are related to those used in our paper, since we also need measures of relaxation. In fact, the work of \cite{gruver2023the} directly inspires one component of our method. On the theoretical side, \cite{PetracheApproximate} studied generalization-approximation tradeoffs in approximately/fully equivariant CNNs in a very general setting, characterizing the effect of model mis-specification on performance. They quantify equivariance as improving the generalization error, and the alignment of data and model symmetries  as improving the approximation error. They leave the impact of improving the optimization error for future work. While we do not provide theoretical results, our work could be seen as focusing on optimization error component of the classical picture\footnote{Characterizing model performance as generalization error $+$ approximation error $+$ optimization error}.

\cite{maile2023equivarianceaware} proposed what they call an \emph{equivariance relaxation morphism}, which reparamterizes an equivariant layer to operate with equivariance constraints on a subgroup, but with the goal of architecture search. \cite{flinth2023optimization} provide an analysis of the optimization dynamics of equivariant models and compare them to non-equivariant models fed with augmented data. However, they don't use the analysis to provide insights on improving the optimization procedure itself. 

Several researchers have recently tried to circumvent optimization difficulties in other ways. For instance, \cite{mondal2023equivariant} suggests using equivariance-promoting canonicalization functions on top of large pre-trained models. The work of ~\cite{BasuSRCVVD23} operates with a similar motivation but without canonicalization. Yet another representative of work with a fine-tuning motivation, but in a different context is \citep{Basu2023EfficientET}. Finally, simplifying equivariant networks with heavy equivariant layers and improving their scalability is an active area of work and is related to easing optimization. Such works usually employ tools from representation theory, tensor algebra, or exploit sampling theorems over compact groups and their homogeneous spaces, such as~\cite{passaro2023reducing, luo2024enabling, cobb2021efficient, ocampo2023scalable}.

\section{Method}\label{sec:method}
To introduce our proposed optimization framework, we first clearly define the equivariant constraint that the models we aim to train must satisfy.
Assume a function $f:\mathbb{R}^n\to \mathbb{R}^m$ and a group $G$~\footnote{We assume henceforth that we are always dealing with Matrix Lie groups.} acting on the input and output spaces via the general linear representations $\rho_\mathrm{in}:G\to \mathrm{GL}(\mathbb{R}^n)$, $\rho_\mathrm{out}:G\to\mathrm{GL}(\mathbb{R}^m)$. Then the function is said to be equivariant to the action of group $G$ if for all $g\in G$ it satisfies the following constraint:
\begin{equation}
        f(\rho_\mathrm{in}(g)x)=\rho_\mathrm{out}(g)f(x),\quad \text{for all }x\in \mathbb{R}^n  \label{eq:Constraint}
\end{equation}
Assuming we use a neural network to approximate the function above, the definition of equivariance as stated doesn't impose specific constraints on the individual layers of the network. Nevertheless, most of the current state-of-the-art equivariant architectures are a composition of simpler layers each one of which is constrained to be equivariant. In this case, the overall model is the result of a composition $f=f_N\circ f_{N-1}\circ\ldots f_2\circ f_1$ of simpler equivariant layers $f_i:V_i\to V_{i+1}$, where $V_i$, $V_{i+1}$ are the input and output spaces on which the group $G$ acts with the corresponding representations $\rho_{V_i}$, $\rho_{V_{i+1}}$ (assuming $V_1=\mathbb{R}^n$, $V_N=\mathbb{R}^m$ are the input and output spaces respectively).  
\par In this work we focus on a family of models as described above---that are defined through a composition of simpler equivariant linear layers. During standard training the linear layers are optimized over the set of intertwiners $H_i$, i.e. the set of linear maps between the representations $(V_i,\rho_{V_i})$ and $(V_{i+1},\rho_{V_{i+1}})$ that have the equivariance property as stated in Equation \ref{eq:Constraint}. The set of intertwiners is only a subset of the set of all possible linear maps from $V_i$ to $V_{i+1}$, and as a result, they have a reduced number of free parameters. We propose to facilitate training over this constrained space by relaxing the constraint imposed on the intermediate layers and optimizing over a larger hypothesis space $\tilde{H}$ which is a superset of the set of equivariant models $H\subset \tilde{H}$. A trivial approach is to completely relax the constraint and solely optimize over the larger set containing all models. The problem with such an approach is that it completely abandons the concept of equivariance and all the attendant generalization benefits. Consequently, to expand the hypothesis space while keeping the benefits of equivariant models, we need a relaxation such that:
\begin{itemize}
    \item Given a non-equivariant model $f\in\tilde{H}$, we can efficiently return to an equivariant one $\bar{f}\in H$.
    \item The relaxed model has a small \emph{equivariance error}   $P_\mathrm{ee}=\mathrm{E}_{x\sim p(x)}\int_G \| \rho_{out}(g)f(x)-f(\rho_{in}(g)x)\|dg.$ This implies that although we extend the space of models we optimize over, we do not diverge too far away from the space of equivariant solutions.
    \item After we project back to the equivariant space, the error of the projection $P_\mathrm{pe}=\mathrm{E}_{x\sim p(x)}\left[\lVert f(x)-\bar{f}(x) \rVert\right]$ is also small. This ensures that while we optimize the less constrained model, we can return to the equivariant one without sacrificing the overall performance.
\end{itemize}
The first  objective can be satisfied by defining an intermediate layer of the form:
\begin{gather}
    f(x)=f_e(x)+\theta Wx,\quad \theta\geq 0\text{ and }f_e\in H, W\in \mathbb{R}^{|V_\mathrm{out}|\times |V_\mathrm{in}|} \label{eq:InitialLayer}
\end{gather}
where $H$ is the set containing all possible equivariant solutions. Here it is easy to see that we can return to an equivariant model by setting $\theta=0$, which we refer to as projection to the equivariant space. The formulation of the linear layer above is similar to the one used in the Residual Pathway Priors (RPP) \citep{finzi2021residual}. Note that in RPP, the value of $\theta$ remains constant and acts as a prior on the level of equivariance we expect from a given task and dataset. Contrasted to that, in this work we aim to control the value of $\theta$ in order to actively change the level of equivariance during training and project back to the equivariance space during inference.

For the second objective, we need to utilize a metric that measures the relative distance of the model from the space of equivariant models $H$. It was observed by \citet{gruver2023the} that an easy way to measure how much a model satisfies the equivariant constraints is by using the norm of the Lie derivative. We present details in the next section.

\subsection{Lie Derivative Regularization Term}\label{sec:liereg}
 Assume we are given a matrix Lie group $G$ acting on a vector space $V$ through its representation $\rho:G\to GL(V)$. For the given group there exists a corresponding Lie algebra $\mathfrak{g}$ with the property that for $A\in \mathfrak{g}$, $e^A\in G$. Additionally, there exists a corresponding Lie algebra representation $d\rho:\mathfrak{g}\to \mathfrak{gl}(V)$ such that $\rho(e^{tA})=e^{d\rho(A)t}$.

If we take the derivative of the action of a group element $e^{tA}\in G$ at $t=0$ we get the Lie derivative:
 \begin{equation}
     \frac{d}{dt}\bigg|_{t=0} \rho(e^{tA})=\frac{d}{dt}\bigg|_{t=0} e^{d\rho(A)t}=d\rho(A)
 \end{equation}
 Assume that the following group representation act on the vector space of functions as:
 \begin{equation}
     \rho_\mathrm{in-out}(g)[f]=\rho_\mathrm{out}(g)^{-1}\circ (f\circ \rho_\mathrm{in}(g))
 \end{equation}
 As observed by \citet{gruver2023the}  the lie derivative of the above action is zero for all equivariant functions $f$, since for all $g\in G$ the action $\rho_\mathrm{in-out}(g)[f]=f$ is the identity map. As a consequence, we can use the norm of the Lie derivative as a metric to compute how much a function $f$ deviates from the equivariant constraint of Equation \ref{eq:Constraint}. For the linear relaxation term  $Wx$ that we introduced in Equation \ref{eq:InitialLayer}, we have that the Lie derivative can be computed in a straightforward manner as:
\begin{align*}
     \mathcal{L}_A(W)=\frac{d}{dt}\bigg|_{t=0} \rho_\mathrm{out}(e^{-At})W\rho_\mathrm{in}(e^{At})&= \frac{d}{dt}\bigg|_{t=0} e^{-d\rho_\mathrm{out}(A)t}We^{d\rho_\mathrm{in}(A)t} \\
     &=-d\rho_\mathrm{out}(A)W+Wd\rho_\mathrm{in}(A)
\end{align*}

As a result, we can measure the degree that a linear layer satisfies the equivariant constraint at a point $x$, by computing the norm of the Lie derivative at that point for each one of the generators of the group. For example in the case where $G$ is the group of 3D rotations ($G=\mathrm{SO}(3)$), we can compute the Lie derivative for each generator:
\begin{align*}
    J_x=\begin{pmatrix}0 & 0 & 0 \\
0 & 0 & -1 \\
0 & 1 & 0\end{pmatrix}, J_y=\begin{pmatrix}0 & 0 & 1 \\
0 & 0 & 0 \\
-1 & 0 & 0\end{pmatrix}, J_z=\begin{pmatrix}0 & -1 & 0 \\
1 & 0 & 0 \\
0 & 0 & 0\end{pmatrix}
\end{align*}
During training, given an input distribution $p(x)$, we compute, for each linear layer $Wx$, the Lie derivative regularization term:
\begin{align*}
    \mathcal{L}_{ld}(W)=\mathrm{E}_{x\in p(x)}\left(\sum_{A\in \{J_i\}} \rVert \mathcal{L}_A(W)x\lVert \right)
\end{align*}
Although the above regularization applies when the symmetry group we are considering is a matrix Lie Group, as we show in the experiments of Section \ref{sec:apprEq}, we can also define a similar regularizer for the case of discrete finite groups. In such a case, for a given linear layer with weights $W$ and input $x$, we compute the sum of the norms of the difference $\mathcal{L}_{g_j}(W)x=\left(\rho(g_j)W-W\rho(g_j)\right)x$
for all generators $g_j$ of the discrete finite group under consideration.

As discussed in \citet{otto2024unifiedframe} and shown in Figure \ref{fig:ldTraining}, the inclusion of the above regularization terms encourages equivariant solutions and prevents the model from diverging away from the space of equivariant models. Moreover, Figure \ref{fig:modelnet} shows how the inclusion of this regularization helps the overall training and results in a performance improvement of the final trained model.

\subsection{Reducing the Projection Error}\label{sec:projError}
While we optimize over a larger hypothesis space, we always aim to return to an equivariant model after the end of training. Using the parametrization in Eq. \ref{eq:InitialLayer} we can always do that by setting $\theta$ to be equal to zero. Although after the projection the resulting model is guaranteed to be equivariant, it might be far from the original relaxed version, meaning it might have a large projection error $P_{pe}$. Specifically, for an individual relaxed layer the projection error is:
\begin{align*}
    P_{pe}=\mathrm{E}_{x\sim p(x)}\left[\lVert f(x)-\bar{f}(x) \rVert\right]&=\mathrm{E}_{x\sim p(x)}\left[\lVert f_e(x)+\theta W x-f_e(x) \rVert\right]\\
    &=\mathrm{E}_{x\sim p(x)}\left[\rVert\theta W x \rVert\right]
\end{align*}
As a result, to ensure that $P_{pe}$ remains low we introduce a second regularization term on the norm $\lVert Wx\rVert$. Additionally, to reduce the contribution of $\theta$ on the projection error, we propose to schedule its value by slowly decreasing it during the last phase of training. Specifically, we apply a cyclic scheduling where given $N_E$ total number of epochs, the value of $\theta$ at epoch $i$ is:
\begin{align*}
    \theta_i=\begin{cases}
        \frac{2i}{N_E} &\text{ if }i<N_E/2\\
        2-\frac{2i}{N_E}&\text{ if }i\geq N_E/2
    \end{cases}
\end{align*}
In Figure \ref{fig:modelnet} we show how both the additional regularization term on the norm of the activation $\lVert Wx \rVert$, and the scheduling of $\theta$, affect the performance of our framework.
\subsection{Training Objective}
Overall, given a task with a corresponding loss $\mathcal{L}_\mathrm{task}$ and a data distribution $D$, our framework optimizes over the following training objective:

\begin{align}
    \mathcal{L}=\mathrm{E}_{(x,y)\sim D}\left[\mathcal{L}_\mathrm{task}(f(x),y)+\lambda_{reg}\sum_{i=1}^N\left(\lVert W_i f_{i-1}(x)\rVert+\sum_{A\in {J_i}}\rVert\mathcal{L}_A(W_i)f_{i-1}(x)\lVert\right)\right]\label{eq:objective}
\end{align}
where $W_i$ is the weight matrix of the $i^\mathrm{th}$ additive unconstrained linear layer and $f_{i-1}(x)$ is the output of the $(i-1)^\mathrm{th}$ layer (with $f_0$ corresponding to the input). 

During training, we control the amplitude of the additive relaxation term by scheduling $\theta$ as described in Section \ref{sec:projError}. During inference, we evaluate only on the equivariant part of the model by setting $\theta=0$. Thus as shown in Figure \ref{fig:method}, after the end of training, the resulting model has the same parameter count and model architecture as the baseline model without any additional additive layers.

\subsection{Relaxing the constraints of different equivariant architectures}\label{sec:relaxconst}
In this section, we consider a selection of different equivariant architectures, and use them to illustrate how we could apply our proposed optimization framework:

\textbf{Vector Neurons \citep{deng2021vn}} In Vector Neurons, the primary linear layer processes features of the form $X\in \mathbb{R}^{N\times 3}$. It achieves equivariance by applying a left multiplication with a weight matrix $f(X)=WX$. This operation mixes only the rows of the input feature matrix and as a result when the input features rotate by $R$, the output also rotates since $f(XR)=WXR=f(X)R$.
\par To apply our proposed relaxation we add a linear layer that allows the mixing of all the elements of the input feature matrix. We can achieve this by unrolling the feature matrix into a vector of dimension $(nm)$ and then after applying an unconstrainted linear layer, roll it back to a feature matrix. So the overall relaxed layer has the form:
\begin{align*}
    f(X)=W_eX+\theta \mathrm{uvec}\left[W\mathrm{vec}(x)\right]
\end{align*}
where $\mathrm{vec}$, $\mathrm{uvec}$ are the corresponding unrolling and rolling operations.

\textbf{SEGNN \citep{brandstetter2021geometric} and Equiformer \citep{liao2023equiformer}}: The intermediate representation of both SEGNN and Equiformer are steerable vector spaces that transform according to a representation of $\mathrm{SO}(3)$. In particular, both models process a collection of type $l$ tensors that transform according to the Wigner-D matrices $D^{(l)}(g)$ of the corresponding type $l$. The interaction between tensors of different types can be done using the Clebsch-Gordan (CG) tensor product which is a bilinear operator that combines two input vectors $x^{l_1}$, $x^{l_2}$ of types $l_1$ and $l_2$ and returns a tensor $(x^{l_1}\otimes x^{l_2})^{l}$ of type $l$ as follows:
\begin{align*}
    (x^{l_1}\otimes x^{l_2})^{l}_m=\sum_{m_1=-l_1}^{l_1}\sum_{m_2=-l_2}^{l_2}C_{(l_1,m_1)(l_2,m_2)}^{(l,m)}x^{(l1)}_{m_1}x^{(l_2)}_{m_2}
\end{align*}
where $x^{(l1)}_{m_1}$, $x^{(l_2)}_{m_2}$ are the $m_1^\mathrm{th}$, $m_2^\mathrm{th}$ elements of tensors $x^{(l1)}$, $x^{(l_2)}$  and $C_{(l_1,m_1)(l_2,m_2)}^{(l,m)}$ are the corresponding CG coefficients. In this operation, the CG coefficients restrict the possible interaction between elements of different types of vectors. We relax the equivariant constraint by adding an unconstrained linear layer that can mix the elements from all the tensors used as intermediate representations, independent of their type. In Equiformer we add such a linear layer in the feed-forward network of the transformer block. Similarly in SEGNN, we add it to the layer that receives the aggregated messages from all the neighbors of a node and updates the node features.

\textbf{Approximately Equivariant Steerable Convolutions \citep{wang2022approximately}}: In this work, the authors designed approximate equivariant steerable convolutional layers. We apply our method by incorporating an additional unconstrained convolutional kernel. Since this task contains discrete symmetry groups, namely discrete rotations and scalings, we replace the Lie derivative regularizer with the corresponding one for discrete groups, described in Section \ref{sec:liereg}.

\section{Experiments}\label{sec:Exp}
\subsection{Equivariant Point Cloud Classification}\label{sec:pcClassification}
We first evaluate our optimization framework by training different networks on the task of point cloud classification.  We use the equivariant variants of PointNet \citep{qi2016pointnet} and DGCNN \citep{dgcnn} which were proposed by \citet{deng2021vn}. We train on the ModelNet40 dataset \citep{shapenet2015}, which contains 12311 point clouds from 40 different classes. We compare with the standard training of these networks using the same hyperparameter configuration as employed in  \citet{deng2021vn}. During both training and testing, we sub-sample the input point clouds to 300 points.

\begin{figure}[t]
    \centering
    \vskip -0.4in
    \includegraphics[width=0.49\textwidth]{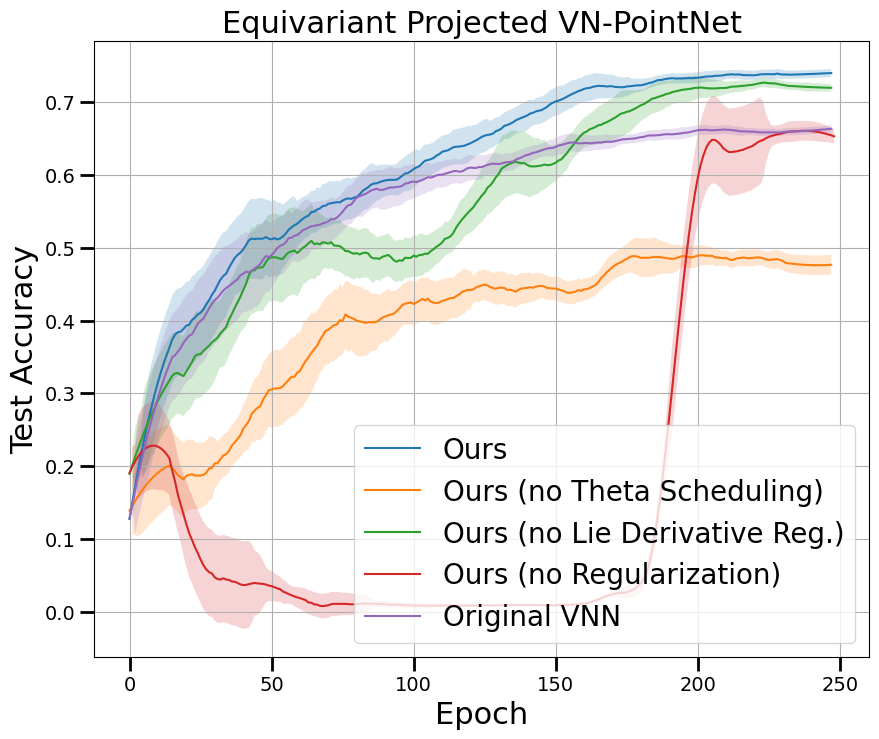}
    \includegraphics[width=0.49\textwidth]{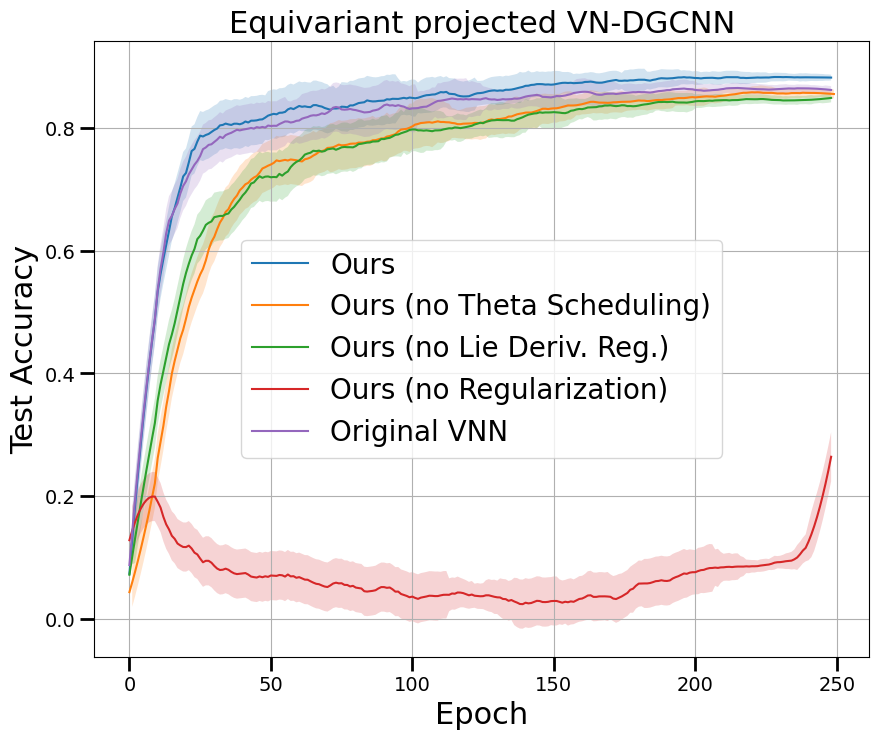}
    \caption{Test accuracy on ModelNet40 classification, during training of equivariant PointNet and DGCNN using a baseline training process and different versions of our method. The accuracy is computed for the equivariant models, i.e. for the models after they are projected in the equivariant space.}
    \label{fig:modelnet}
\end{figure}

To apply our method we relax the Vector Neurons linear layer by following the methodology described in Section \ref{sec:relaxconst}. For both networks we set the regularization term $\lambda_{reg}=0.01$, which is a value we use in all of the following experiments. We provide a more detailed description of the training parameters in Appendix \ref{sec:traindet}. Furthermore, in Appendix \ref{sec:crossval} we  describe the process of choosing the hyperparameter $\lambda_{reg}$ and show the method's robustness to its value. Figure \ref{fig:modelnet} showcases how applying our proposed framework benefits the training of both networks. Specifically, for the case of the smaller and less performant PointNet, we can see an even larger performance increase over the baseline. These results show how the performance benefits of our optimization framework increase in smaller under-parametrized networks, an effect we investigate further in Section \ref{sec:scaling}. In Appendix \ref{sec:compOverhead} we provide additional details on the computational and memory overhead of our proposed optimization, showcasing that while additional parameters are introduced during training the overhead in the training time is limited.

\begin{figure}[t]
    \centering
    \vskip -0.10in
    \subfigure[]{
        \includegraphics[width=0.55\textwidth]{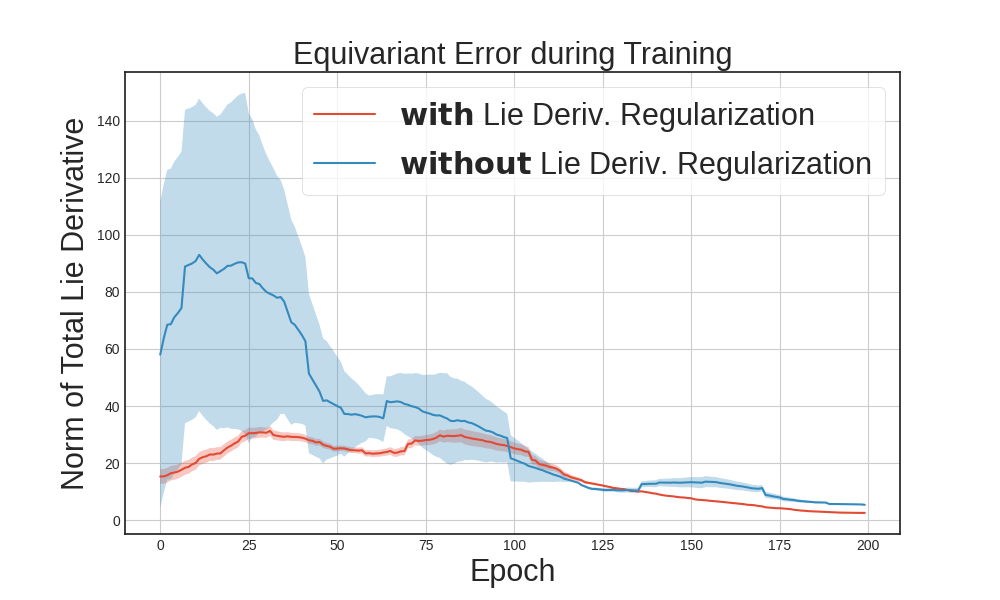}\label{fig:ldTraining}}
    \subfigure[]{
        \includegraphics[width=0.42\textwidth]{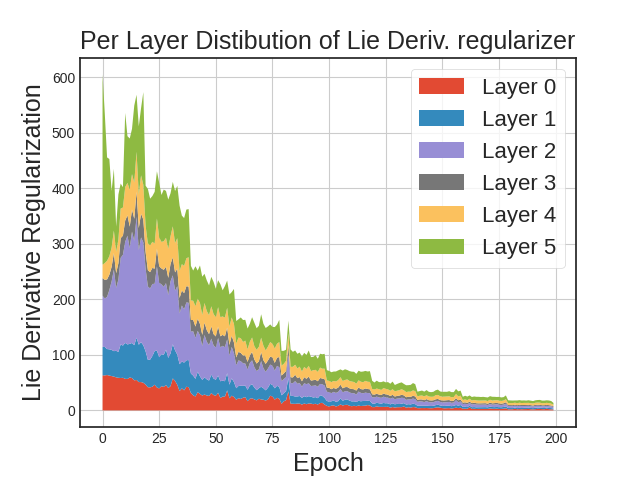}}
    \label{fig:lieDerivePlots}
    \caption{(a) Norm of the total Lie derivative of the relaxed PointNet model trained with and without the Lie derivative regularization term. For the computation of the Lie derivative we use the method proposed in \cite{gruver2023the}. (b) Value of the Lie derivative regularization term for each individual layer of the relaxed PointNet model while we train using our framework and with Lie derivative regularization weight set to $\lambda_\mathrm{reg}=0.01$}
\end{figure}

\textbf{Ablations on the regularization terms and $\theta$ scheduling}: In addition to the training curves of our method and of the baseline, Figure
\ref{fig:modelnet} shows the accuracy of our proposed optimization procedure when we remove some of the proposed regularization terms or the scheduling of $\theta$. We observe that without any regularization both models diverge from the space of equivariant solutions. As a result, during inference when $\theta=0$ their projection error $P_{pe}$ becomes larger, resulting in a significant drop in test accuracy. Similarly,  without the Lie derivative regularizer, the final test accuracy of both network variants drops. In such cases, $Wx$ is unconstrained and can learn to extract non-equivariant features that the equivariant part $f_e$ is not able to learn in any stage of the training. This effect can also be observed in figure \ref{fig:ldTraining} showing the total Lie derivative of the network when it is trained with and without the lie derivative regularization term. Not including the Lie derivative regularization allows the network, especially in the beginning of training, to optimize over solutions with large equivariance error. Finally, for both networks, we observe that $\theta$ scheduling, as described in Section \ref{sec:projError}, can benefit training compared to fixing $\theta$ to a constant low value. In Appendix \ref{sec:thetaAblation} we provide additional results showcasing how contrary to our method  a model with a constant $\theta$ (without $\theta$ scheduling) has a significant drop in performance after it is projected into the equivariant space.
\subsection{Scaling on Different Model and Dataset Sizes}\label{sec:scaling}
To better understand how the model and dataset sizes affect our proposed optimization framework, we train models of variable depth on different numbers of training samples. As a baseline model we use the Steerable E(3) GNN (SEGNN) \citep{brandstetter2021geometric} and we train it on the task of Nbody particle simulation \citep{pmlr-v80-kipf18a}. This task consists of predicting the position of 5 electrically charged particles after 1000 time steps when given as input their initial positions, velocities, and charges.

\begin{figure}
    \centering
    \vskip -0.1in
    \subfigure[]{
    \includegraphics[width=0.48\textwidth]{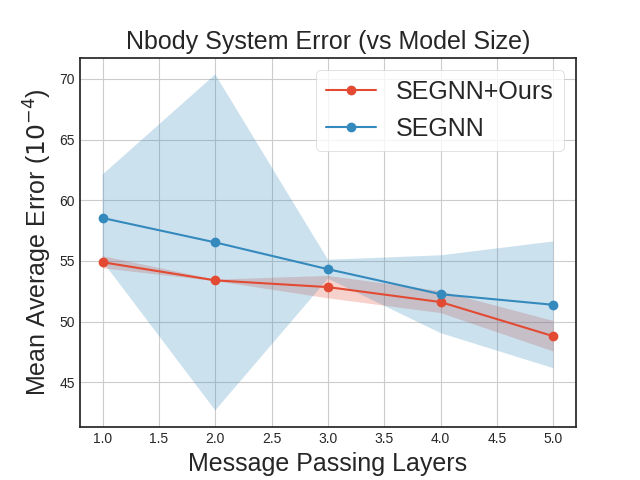}\label{fig:errorVsModelSize}}
    \subfigure[]{
    \includegraphics[width=0.48\textwidth]{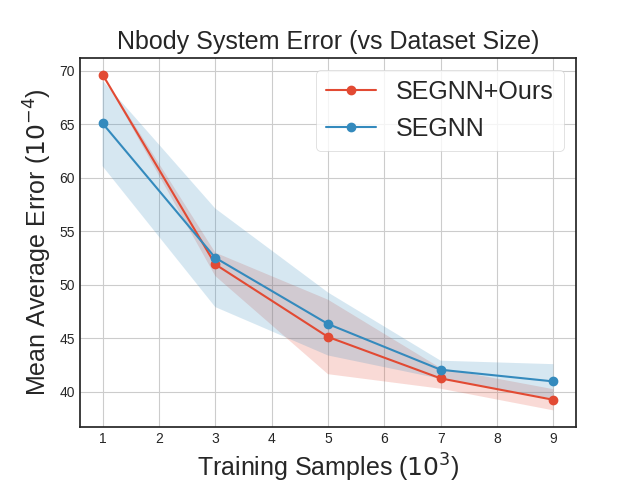}\label{fig:errorVsDatasetSie}}
    \caption{Mean Average Error on the Nbody particle simulation for (a) different model sizes, (b): different dataset sizes.}
    \label{fig:enter-label}
\end{figure}

Figure \ref{fig:errorVsModelSize} shows the mean average test error achieved by networks of different sizes, both when trained with a standard optimization, and when trained with our proposed framework. We can observe that for all sizes our method achieves better generalizations. The gap between our method and the baseline becomes greater in the cases of smaller networks, a phenomenon that we also observed in the point cloud classification experiments in Section \ref{sec:pcClassification}.  Thus, our framework, by relaxing the constraint and introducing additional degrees of freedom, can help the overall optimization, especially in models with a limited number of parameters. Additionally, figure \ref{fig:errorVsDatasetSie}  shows that when we fix the model size and increase the dataset size our method is able to scale better than the baseline. In both cases, we can observe that the training of the baseline has a much larger variance and is highly dependent on the random initialization of the layers. On the contrary, our method results in a more consistent training with a smaller variance between the random trials.
\subsection{Molecular Dynamics Simulation}\label{sec:moldynamics}
To evaluate our framework in a challenging task using a complex network architecture, we train Equiformer \citep{liao2023equiformer} on the task of molecular dynamics simulations for a set of molecules provided as part of the MD17 dataset \citep{md17}. The goal of this task is to predict the energy and forces from different configurations of a pre-specified molecule. Following \citet{liao2023equiformer}, for each molecule we use only 950 different configurations for training which significantly increases the task difficulty. For all training runs we use the same value of $\lambda_{reg}=0.01$ as in the previous experiments and for the rest of the hyperparameters, we use the same configuration as the one proposed in \citet{liao2023equiformer}. In Table \ref{tab:equiformer_selected_molecules} we show that the mean absolute error of energy and force prediction achieved by Equiformer, both when it is trained using standard training, and when it is trained with our proposed optimization framework. Without any additional hyperparameter tuning, our framework is able to provide improvements on the performance of Equiformer even for this challenging data-scarce task.
\begin{table}[ht]
\centering
\caption{MAE of Equiformer trained with and without our optimization framework on a set of molecules from the MD17 dataset. The energy is reported in meV and the force in meV/{\AA} units}
\label{tab:equiformer_selected_molecules}
\begin{tabular}{@{}lcccccccc@{}}
\toprule
& \multicolumn{2}{c}{Aspirin} & \multicolumn{2}{c}{Benzene} & \multicolumn{2}{c}{Ethanol} & \multicolumn{2}{c}{Salicylic acid} \\
\cmidrule(r){2-3} \cmidrule(r){4-5} \cmidrule(r){6-7} \cmidrule(r){8-9}
Methods & Energy & Forces & Energy & Forces & Energy & Forces & Energy & Forces \\
\midrule
Equiformer  & 5.3 & 7.2 & 2.2 & 6.6 & 2.2 & 3.1 & 4.5 & 4.1 \\
Equiformer+Ours & \textbf{5.2} & \textbf{7.1} & 2.2 & 6.6 & \textbf{2.0} & \textbf{2.9} & \textbf{4.1} & 4.1 \\
\bottomrule
\end{tabular}
\end{table}
\subsection{Optimizing Approximately Equivariant Networks}\label{sec:apprEq}
Finally, we show how our framework can be beneficial, not only for the optimization of exactly equivariant networks, but also for approximate equivariant ones. We apply our method on top of the approximate equivariant steerable CNNs proposed in \citet{wang2022approximately}. Although these models are not projected back to the  equivariant space, they are still regularized to stay within solutions with small equivariant error. The main difference with our framework is that in the approximate equivariant setting, the equivariant relaxation remains the same throughout training. On the contrary, we propose to progressively constrain the model by modulating the value of the unconstrained term by slowly decreasing the value of $\theta$ from Equation \ref{eq:InitialLayer}. As a result by applying our optimization framework on top of the standard training of the approximate equivariant kernels, we test how progressively introducing additional constraints throughout training can help the performance of the network. 

We evaluate our method on the task of 2D smoke flow prediction described in \citet{wang2022approximately}. The inputs of the model are sequences of successive $64\times 64$ crops of a smoke simulation generated by PhiFlow \citep{phiflow}. The desired output is a prediction of the velocity field for the next time step. We evaluate in two different settings: the "Future" setting where we evaluate on the same part of the simulation but we predict future time steps not included in the training, and the "Domain" setting where we evaluate on the same time steps as in training but in different spatial locations. The data are collected from simulations with different inflow positions and buoyant forces. For the rotational symmetry case, while the direction of the inflow and of the buoyant force is symmetric to $90^\circ$ degrees rotations ($C_4$ symmetry group), the buoyancy factor changes for the different directions making the task not symmetric. For the scaling symmetry, the simulations are generated with different spatial and temporal steps, with the buoyant factor changing across scales.

In addition to the approximate equivariant steerable CNNs (RSteer), we compare with a simple MLP, with a non-equivariant convolutional network (ConvNet), as well as with an equivariant convolutional network (Equiv) \citep{e2cnn}  and with two additional approximate equivariant networks RPP \citep{finzi2021residual} and LIFT \citep{wang2021equivariant} that are trained using a standard training procedure. In Table \ref{tab:approximate} we see that by applying our optimization framework the resulting approximate equivariant model outperforms all other baselines in both cases of approximate rotational and scale symmetry. These results indicate that starting from an unconstrained model and progressively increasing the applied constraints can benefit optimization even in the case where at the end of training we stay in the space of approximate equivariant models and do not project back to the equivariant space.
\begin{table}
\centering
\caption{RMSE error on the synthetic smoke plume dataset with approximate rotational and scale symmetries. In the "Future" evaluation we train and evaluate the models in the same simulation location but we test for later time steps in the simulation from the ones used in training. In the "Domain" evaluation we train and evaluate the models on the same timesteps but on different spatial locations in the simulation.}
\label{tab:approximate}
\resizebox*{\textwidth}{!}{
\begin{tabular}{cc|cccccc|c}
\hline
\multicolumn{2}{c}{\textbf{Model}} & \textbf{MLP} & \textbf{Conv} & \textbf{Equiv} & \textbf{Rpp} & \textbf{Lift}  & \textbf{RSteer}& \textbf{RSteer+Ours} \\ \hline
\multirow{2}{*}{Rotation}&Future & $1.38\pm0.06$ & $1.21\pm0.01$ & $1.05\pm0.06$ & $0.96\pm0.10$ & $0.82\pm0.08$ &  $0.80\pm0.00$ & $\bm{0.79\pm 0.01}$ \\ 
&Domain & $1.34\pm0.03$ & $1.10\pm0.05$ & $0.76\pm0.02$ & $0.83\pm0.01$ & $0.68\pm0.09$ &  $0.67\pm0.01$ & $\bm{0.58\pm 0.00}$ \\ \hline
\multirow{2}{*}{Scale}&Future & $2.40\pm0.02$ & $0.83\pm0.01$ & $0.75\pm0.03$ & $0.81\pm0.09$ & $0.85\pm0.01$ &  $0.70\pm0.01$ & $\bm{0.62\pm 0.02}$ \\ 
&Domain & $1.81\pm0.18$ & $0.95\pm0.02$ & $0.87\pm0.02$ & $0.86\pm0.05$ & $0.77\pm0.02$ &  $0.73\pm0.01$ & $\bm{0.67\pm 0.01}$ \\ \hline
\end{tabular}}
\end{table}
\section{Conclusion}
In this work, we focus on the optimization of equivariant NNs. We proposed a framework for improving the overall optimization of such networks by relaxing the equivariance constraint and optimizing over a larger space of approximately equivariant models. We showcase the importance of utilizing regularization during training to ensure that the relaxed models stay close to the space of equivariant solutions. After training, we project back to the equivariant space arriving at a model that respects the data symmetries, while retaining its high performance on the task. We evaluate our proposed framework and its individual components over a variety of different equivariant network architectures and training tasks, and we report that it can consistently provide performance benefits over the standard training procedure. A theoretical analysis of our approach, possibly with appeal to empirical process theory~\citep{pollard1990empirical} to control the optimization error, is left for future work. 
\section*{Acknowledgements}
We gratefully acknowledge support by the following grants: NSF FRR 2220868, NSF IIS-RI 2212433,
ARO MURI W911NF-20-1-0080, and ONR N00014-22-1-2677.
\bibliography{main}
\bibliographystyle{main}
\newpage
\appendix
\section{Appendix/ Supplemental Material}\label{sec:supMaterial}
\subsection{Training Details}\label{sec:traindet}
In this section, we provide additional details for the application of our framework in the experiments presented in this work. We fix the weight of the regularization term to be $\lambda_{reg}=0.01$ for all the experiments. We arrive that the above value for the hyperparameter $\lambda_{reg}$ by performing grid-search using cross validation, as described in more detail in Section \ref{sec:crossval}.
Additionally, except on the corresponding  ablation study, we use the scheduling of the value of $\theta$ as described in Section \ref{sec:projError}. The variance reported is over 5 random trials of the same experiment with different seeds. We run all the experiments on NVIDIA A40 GPUs. For the model-specific training details:
\begin{itemize}
    \item \textbf{Point Cloud Classification}: We use as baselines the VN-PointNet and VN-DGCNN network architectures described in the work of \cite{deng2021vn}. For the relaxed version of VN-PoitNet we train for 250 epochs using the Adam optimizer \citep{kingma2015adam}, with an initial learning rate of $10^{-3}$, that we decrease every 20 epochs by a factor of 0.7, and weight decay equal to $10^{-4}$. For the relaxed version of VN-DGCNN we train for 250 epochs using stochastic gradient descent, with an initial learning rate of $10^{-1}$, that we decrease using cosine annealing, and weight decay equal to $10^{-4}$. The batch size used was 32.
    \item \textbf{Nbody particle simulation}: We train the relaxed version of SEGNN \citep{brandstetter2021geometric} for 1000 epochs using Adam optimizer \citep{kingma2015adam} with a learning rate of $5*10^{-4}$,  a weight decay of $10^{-12}$ and batch size of 100. We report the test MAE for the model at the training epoch that achieved the minimum validation error.
    \item \textbf{Molecular Dynamics Simulation}: We train the relaxed version of Equiformer \citep{liao2023equiformer} for 1500 epochs using AdamW optimizer \citep{loshchilov2018decoupled} with an initial learning rate of $10^{-5}$, that we decrease using cosine annealing, and with weight decay equal to $10^{-6}$. The batch size used was 8. We use the network variant with max representation type set to $L_\mathrm{max}=2$
    \item \textbf{Approximately Equivariant Steerable Convolution}: We train the approximately equivariant steerable convolution proposed in \cite{wang2022approximately} after we apply our additional relaxation. We modify the same architecture used in the original work which contains 5 layers of approximate equivariant steerable convolutions. We train for 1000 epochs using the Adam optimizer \citep{kingma2015adam}. We use an initial learning rate of $10^{-4}$, that we decrease at each epoch by $0.95$. We perform early stopping, where the stopping criterion is that the mean validation score of the last 5 epochs exceeds the mean validation score of the previous 5 epochs.
\end{itemize}
\subsection{Choice of Hyperparameters}\label{sec:crossval}
\begin{figure}
    \centering
    \includegraphics[width=0.5\linewidth]{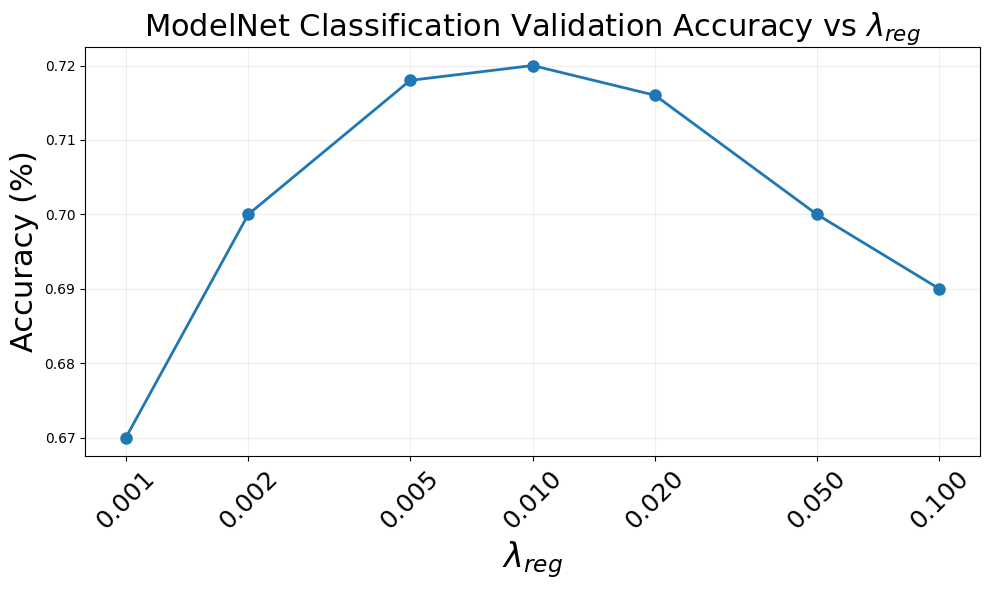}
    \caption{ModelNet40 classification accuracy on the validation set using our proposed method with different values of $\lambda_{reg}$. The base model used was the VN-PointNet. The model was trained on a split of the training set containing 80\% of the training data. The other 20\% of the data were held out as the validation set used to evaluate the model.  }
    \label{fig:hypersearch}
\end{figure}

In all the experiments, apart from the weight $\lambda_{reg}$ of the proposed regularization term, we use the same hyperparameters as the ones used by the baseline methods we compare with. For the choice of $\lambda_{reg}$ we perform hyperparameter grid search using cross-validation with an 80\%-20\% split of the original training set of ModelNet40 into training and validation. Figure \ref{fig:hypersearch} showcases the performance of a VN-Pointnet model trained with our method on the 80\% training split and evaluated on the 20\% validation split for different values of $\lambda_{reg}$. We observed that the best value of $\lambda_{reg}$ is relatively robust across tasks, so we performed an extensive hyperparameter search for the task of point cloud classification, and we used the found value $\lambda_{reg}=0.01$ across all other tasks.
\subsection{Computational and Memory Overhead of proposed method}\label{sec:compOverhead}
The computational overhead introduced by our method only affects the training process. During inference, after the projection to the equivariant space, the retrieved model has the same architecture as the corresponding base model to which we apply our method on, thus it also has the same computational and memory requirements. In Table \ref{tab:comp_overhead} we show the cost of our method, in terms of training time and number of additional parameters. While our proposed method introduces additional parameters during training, due to the parallel nature of the unconstrained non-equivariant term, the overhead in training time can be limited given enough memory resources. As a result, while the additional parameters are approximately three times the parameters of the base model the increase in the training time is below 10\% of the base training time.
\begin{table}[ht]
\centering
\caption{Additional Number of parameters and Computational Overhead introduced by the proposed method}
\resizebox{\textwidth}{!}{%
\begin{tabular}{|l|c|c|c|c|}
\hline
\multirow{2}{*}{\textbf{Model}} & \textbf{Number of Parameters} & \textbf{Additional Parameters} & \textbf{Time per Epoch} & \textbf{Time per Epoch} \\ 
& \textbf{(Base Model)} & \textbf{(Ours)} & \textbf{(Base Model)} & \textbf{(Ours)} \\ \hline
VN-PointNet & 1.9M & 6.4M  & 75s & 80s \\ \hline
VN-DGCNN    & 1.8M & 6.2M  & 148s & 154s \\ \hline
Equiformer & 3.4M & 10M & 52s & 57s \\ \hline
\end{tabular}}
\label{tab:comp_overhead}
\end{table}
\subsection{Ablation on $\theta$ Scheduling and Equivariant projection }\label{sec:thetaAblation}
\begin{figure}
    \centering
    \includegraphics[width=0.5\linewidth]{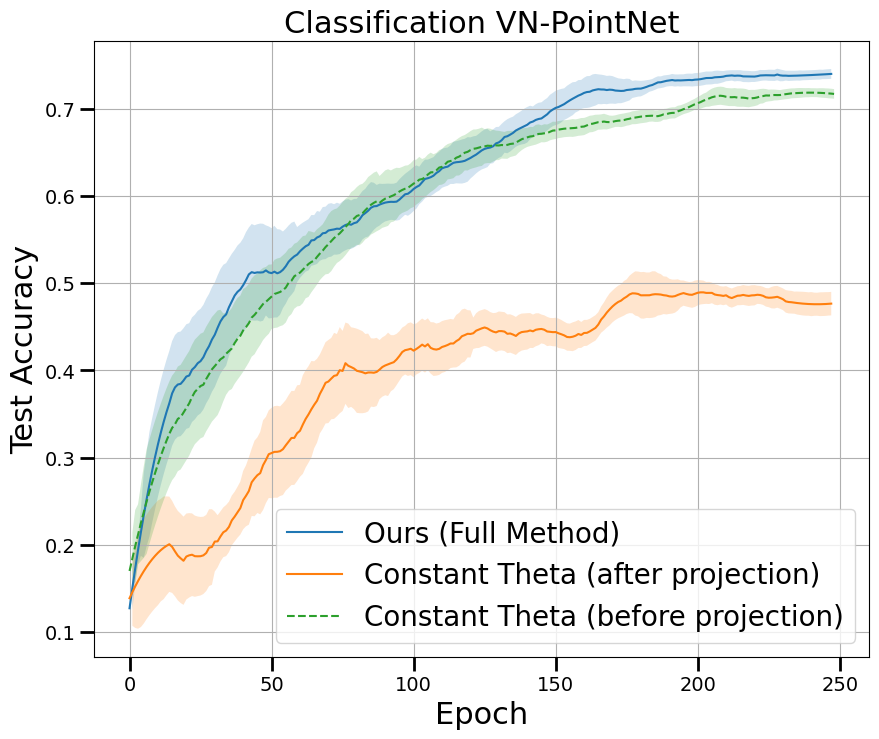}
    \caption{Comparison of the performance on ModelNet40 classification (300 points), for a model trained with our method and a model trained with a constant value of $\theta$, for which the level of equivariant error is controlled only by the regularization term. For the latter method we show results both before and after the projection to the equivariant space.}
    \label{fig:ablationProject}
\end{figure}

During the later stages of training our proposed $\theta$ scheduling slowly decreases the level of relaxation of the equivariant constraint, bringing the model closer to the equivariant space. This process allows the model to smoothly transition from the relaxed equivariant case to the exact equivariant one. In Figure \ref{fig:ablationProject} we show a comparison of the performance of a model trained with our proposed $\theta$ scheduling and a model trained with a constant $\theta$ before and after it is projected to the equivariant space. While the performance of the relaxed equivariant model with constant $\theta$ is close to the performance achieved by our method, we can observe a sudden drop in performance once it is projected back to the equivariant space. On the other hand, our proposed scheduling of $\theta$ allows the model to return to the equivariant space by the end of training without showcasing such a significant performance drop.

\subsection{Additional comparison with Methods using Equivariant Adaptation/Fine Tuning}
As described in Section \ref{sec:related}, while our work focuses on improving the optimization of equivariant networks, the works of \citet{mondal2023equivariant} (Equiv-Adapt) and \citet{BasuSRCVVD23} (Equi-Tuning) focus on circumventing the need of optimizing equivariant network by performing equivariant adaptation or fine tunining of non-equivariant models. Since such methods have a different focus, mainly utilizing already pre-trained non-equivariant models to solve equivariant tasks, a straightforward comparison can be challenging. Nevertheless, in Table \ref{tab:equivariantAdapt} we provide a comparison with our proposed method on the task of sparse point cloud classification. We can see that our method outperforms Equiv-Adapt and has performance close to the one achieved by Equi-Tuning which requires multiple forward passes during inference. 

\begin{table}[ht]
\centering
\caption{Comparison of our proposed method with previous works performing equivariant adaption or finetuning, on ModelNet40 classification (Base model: VN-PointNet). Here it is important to note that in the case of Equi-Tuning, equivariance is achieved by group averaging. As a result, during inference the model is required to perform multiple forward passes, which slows down the method's inference.}
\begin{tabular}{|l|c|c||c|}
\hline
     Equiv-Adapt & Equi-Tuning & Original VNN & \textbf{Ours} \\ \hline
     66.3\% & 74.9\% & 66.4\% & 74.5\% \\ \hline
\end{tabular}
\label{tab:equivariantAdapt}
\end{table}

\subsection{Limitations}\label{sec:limitations}
As we describe in Section \ref{sec:method}, our work focuses on the assumption that the equivariant NN satisfies the equivariant constraint by imposing it in all of its intermediate layers. Although this assumption is general enough to cover a large range of state-of-the-art equivariant architecture, it doesn't apply to all possible equivariant networks since it is possible to ensure overall constraint satisfaction using a different methodology. Additionally the proposed regularizers in Section \ref{sec:liereg} can be applied to tasks where the symmetry group is a matrix Lie group or a discrete finite group. Extending our proposed framework to arbitrary symmetry groups is a future research question that is not addressed in this paper.
\subsection{Broader Impact}\label{sec:broaderImpact}
This paper focuses on the question of improving the optimization of equivariant neural networks. Such equivariant networks are currently used to solve tasks in different fields-- ranging from computer vision to computational chemistry. As a result, its broader societal impact is highly dependent on the specific network it enables optimizing.  

\end{document}